\documentclass{article}

\usepackage{PRIMEarxiv}
\usepackage[numbers]{natbib}
\usepackage[utf8]{inputenc} 
\usepackage[T1]{fontenc}    
\usepackage{hyperref}       
\usepackage{url}            
\usepackage{booktabs}       
\usepackage{amsfonts}       
\usepackage{nicefrac}       
\usepackage{microtype}      
\usepackage{lipsum}
\usepackage{fancyhdr}       
\usepackage{graphicx}       

\title{Extracting triples from dialogues for conversational social agents

}

\author{
  Piek Vossen, Selene Báez Santamaría, Lenka Bajčetić, and Thomas Belluci \\
  Computational Linguistics and Text Mining Lab, \\
  Vrije Universiteit Amsterdam \\
  Amsterdam\\
  \texttt{\{s.baezsantamaria, piek.vossen\}@vu.nl} \\
  \texttt{lenka.bajcetic@ic.etf.bg.ac} \\
}

\begin{document}
\maketitle

\begin{abstract}
Obtaining an explicit understanding of communication within a Hybrid Intelligence collaboration is essential to create controllable and transparent agents. In this paper, we describe a number of Natural Language Understanding models that extract explicit symbolic triples from social conversation. Triple extraction has mostly been developed and tested for Knowledge Base Completion using Wikipedia text and data for training and testing. However, social conversation is very different as a genre in which interlocutors exchange information in sequences of utterances that involve statements, questions, and answers. Phenomena such as co-reference, ellipsis, coordination, and implicit and explicit negation or confirmation are more prominent in conversation than in Wikipedia text. We therefore describe an attempt to fill this gap by releasing data sets for training and testing triple extraction from social conversation. We also created five triple extraction models and tested them in our evaluation data. The highest precision is 51.14 for complete triples and 69.32 for triple elements when tested on single utterances. However, scores for conversational triples that span multiple turns are much lower, showing that extracting knowledge from true conversational data is much more challenging.
\end{abstract}


\section{Introduction}
In a Hybrid Intelligence context in which humans and agents need to collaborate, social information exchange is crucial. Recently, generative Large Language Models (LLMs) \cite{ouyang2022training, touvron2023llama} excel in having social conversations with people; yet the knowledge exchanged remains implicit. Our goal is to develop conversational agents that explicitly represent the information exchanged in a way that can function as an episodic memory, for example, as Knowledge Graphs (KGs)~\cite{baez-santamaria-2024-knowledge}. This will give an agent more control over the goals of the interaction and allow for explicit symbolic reasoning. Especially in collaborative scenarios, such an explicit representation can provide transparency and explainability. In order for social conversational agents to create such KGs, they need to detect information and perspectives expressed in dialogue as RDF\footnote{Resource Description Framework: \url{https://www.w3.org/RDF/}} triples that can be added to the memory graph.

People often talk about each other in social conversations. Most conversations are therefore homodiegetic \cite{eisenberg2020automatic}, telling a story from the perspective of the speaker. It is not surprising that the most frequent subjects and objects in conversational utterances are the pronouns "I" and "you" referring to the speakers \cite{santamaria2023role}. However, little is known about the information that is exchanged during such conversations. Hence, we also do not know what conversational social agents need to understand from such conversations. Most of the state-of-the-art technology for extracting explicit knowledge concentrates on Wikipedia texts, primarily because of resource availability. 

However, social conversation contains a mixture of facts and personal perspectives \cite{santamaria2023role} that are expressed through sequential dialogue turns between interlocutors. As partial information is provided by different speakers, it is important to deal with coreference across these utterances, ellipsis, coordination, implicit and explicit negation, as well as speaker perspectives such as certainty, judgments, and emotions. This makes it challenging to develop conversational agents that can have social conversations with humans and extract the explicit knowledge that is provided. 

In this work, we focus on the knowledge exchanged in social conversation. We define training and test data for the task of triple extraction. We compare several models, ranging from dedicated context-free-grammars, dependency parsing, to fine-tuned and prompted Large Language Models (LLMs). 

Our contributions are as follows.

\begin{itemize}
    \item An English test suite for dialogue turns annotated with social facts and perspectives as triples, categorized by dialogue acts and types of knowledge.
    \item Training data for dialogue turn sequences extracted from PersonaChat, DailyDialog, and Circa annotated with fact triples and perspectives.
    \item Five models for the triple extraction of social conversations.
\end{itemize}

Our data and models are freely available on GitHub\footnote{\url{https://github.com/leolani/cltl-knowledgeextraction}}

\section{Related work}
\label{sec:related-work}
Information extraction is the task of identifying factual information in unstructured data and representing this knowledge in a structured form. Knowledge graphs (KG) are one of such forms, where nodes represent entities and edges represent relationships between these entities. In that framework, a fact is then represented as a triple: \textit{subject} - \textit{predicate} - \textit{object}~\cite{reinanda2020knowledge}.

\paragraph{Types of information extraction}
Information extraction may be divided into two categories: Closed Information Extraction (CIE) and Open Information Extraction (OIE). On the one hand, CIE aligns the extracted information with existing entities, relations, or attributes in a predefined KG. This approach relies on (semi-)supervised methods to extract triples, often combining Named Entity Recognition, Entity Linking, and Relation Extraction. CIE systems are typically tailored to a specific domain and KG schema, such as SKOS or DBpedia~\cite{luan-etal-2019-general}. Therefore, the KG schema determines the set of recognizable entities and relations a priori.

In contrast, OIE is not limited to a predefined set of entities and relations. Instead, it involves extracting spans from text that represent entities and their relationships without prior alignment to a KG. This open nature makes OIE suitable for extracting new or domain-independent knowledge.

\paragraph{Historical perspective}
Information extraction is a well-established research area that has evolved significantly over time. The early work was manual-intensive and rule-based methods were developed primarily out of linguistic insight\cite{hearst-1992-automatic}. Most of these rule-based systems focused on extracting is-a relations, thus forming taxonomies in specialized domains. Over time, the field has shifted to feature-based systems, with the emergence of OIE. Recent advances in deep learning have introduced powerful architectures capable of tackling subtasks like NER and RE with greater generality.

Despite advances, relation extraction remains a significant bottleneck, often yielding lower accuracy compared to NER. Developing robust general-purpose relation extractors continues to be an open challenge in the field~\cite{huang-wang-2017-deep}.

\paragraph{Extracting personal knowledge}
The task of DialogNLI was introduced by Welleck et al. \cite{welleck2019dialogue} to determine whether pairs of utterances from social conversations entail, contradict or are unrelated to one another. For this purpose, they annotated single sentences from the PersonaChat dataset \cite{zhang2018personalizing} with triples and possibly negation values, e.g., <I, like\_drink, espresso>. However, Welleck et al. did not train a triple extractor from this data but used the triple annotations to train a classifier to infer the NLI relationship between turns represented through these triples. Wang et al. \cite{wang-etal-2022-extracting}, on the other hand, did use the annotated DialogNLI data to fine-tune GPT-2 to extract structured information about a person, such as their hobbies, pets, family, likes and dislikes, which we define here as \textbf{social facts}, in the form of triples. Their approach comes closest to our work, but whereas they consider single turns to extract implicit properties, we consider both single turns and conversational sequences of turns that are more complex. Furthermore, they propose a single generative model, and we experimented with various smaller models and produced a new data set both for turn-level and conversation-level extraction.



\section{Conversational information exchange}
\label{sec:conversations}

The extraction of world knowledge and personal information from dialogue presents unique challenges from a linguistic point of view. Although alternative domains, such as news and Wikipedia articles, can be presumed to consist of complete grammatical sentences, dialogues are often highly fragmented and can give rise to many kinds of ungrammatical non-sentential constructions \cite{jurafskyspeech, schlangen2003interpretation}. Dialogue is defined as a joint activity involving two or more participants in which information or ideas are exchanged through natural spoken or written language \cite{fernandez2007classifying, jurafskyspeech}. In this view, the primary purpose of dialogue is simply to provide a means to share information, build common ground, and allow one to learn about others, their beliefs, and the world around them. In particular, the perspectives of the interlocutors on social facts as defined in the GRaSP model \cite{van2016grasp, fokkens2017grasp, vossen2022grasp} add another layer of complexity, both in terms of language structure and status of information. Interlocutors can express their epistemic beliefs (confirm or denial), their feelings and emotions, and the certainty of any belief using more complex linguistic constructs. For example, "I think it could be nice to stay at home and rest instead of going to work" expresses 1) uncertainty, 2) emotion, and 3) the claims of (a) staying at home to rest, and (b) not to go to work; all in one utterance.

In the following examples, we highlight several noteworthy dialogue phenomena including ellipsis, the use of anaphora, and the speaker's perspective. Consider the following example dialogues taken from the DailyDialog data set.

\paragraph{Explicit Negation}
In the next example, "Speaker 2" asks a direct question to "Speaker 1" and receives a negative answer (polarity=-1) with a negative sentiment expressed (-1), showing different perspective values for the claim of having kids. We see here that social facts and perspectives are mixed and need to be inferred from multiple turns across interlocutors.


\begin{itemize}
    \item Explicit-no  \label{ex:explicit-no}:
    \begin{description}
    \item[Speaker 1:] Ow wow now that will be hard task
    \item[Speaker 2:] Yeah do you have kids?
    \item[Speaker 1:] No, I sadly do not. I never been one to hold a steady relationship
    \item[triple:] <speaker1, have, kids>
    \item[perspective:] polarity=-1, certainty=01, sentiment=-1
    \end{description}
\end{itemize}

\paragraph{Implicit Negation}
The two examples below 
are more complex, showing two dialogues with implicit negation. In the first, "Speaker 1" states that they do not have time to watch a new drama, which only indirectly negates watching it (polarity=-1) with high, not absolute, certainty (0.8). The sentiment is still positive (1) as we can assume that "Speaker 1" would like to see it. In the second fragment, "Speaker 1" does not directly answer the question of "Speaker 2" but makes another statement that makes it unlikely that "Speaker 1" likes living there. In this case the polarity is also negative, but certainty can be expected to be lower, and, in general, "Speaker 1" expresses negative sentiment towards living there. 

\begin{itemize}
\item Implicit-no \label{ex:implicit-no}:
    \begin{description}
    \item[Speaker 1:] I do work! I waitress during the day, then I do side jobs during the evening.
    \item[Speaker 2:] Will you be watching that new drama this evening?
    \item[Speaker 1:] I do not have any time to start a new series today
    \item[triple:] <speaker1, watch, new\_drama>
    \item[perspective:] polarity=-1, certainty=0.8, sentiment=1
    \end{description}
\item Implicit-no, coreference, coordination  \label{ex:implicit-no-coref-coord}:
    \begin{description}
    \item[Speaker 1:] Yup been to new york city 3 times this year
    \item[Speaker 2:] Do you want to live in New York city?
    \item[Speaker 1:] The city is too busy and loud
    \item [triple:] <speaker1, like, living\_in\_NY>
    \item[perspective:] polarity=-1, certainty=0.5, sentiment=-1
    \end{description}
\end{itemize}

Note that triple elements are spread over the different turns, such as "watching" and "like living\_in\_NY" in turn 2 of both conversations. We also observe coreference of "the city" with "New York city" and coordination of two claims about NY, being both too busy and too loud.

\vspace{.5cm}
To capture phenomena such as the above, we created two data sets with conversational utterances. The social facts expressed are annotated as triples, along with the perspective information of the sources. We first describe a test suite with single turns and, second, annotations of existing conversational data between crowdworkers annotated in a similar way over sequences of turns.

\subsection{Test suite of turns}
\label{lab:turn-level-data}
We built a dedicated dataset with single-turn statements and questions given human and agent speaker roles. The data were specifically designed to capture distinct domains and linguistic phenomena. For example, we created datasets for open verb questions and WH-questions separately. This diverse collection allowed us to test the capabilities of each extractor on a wider range of conversational topics. In total, nine manually curated test sets were created, where each test item consists of the text and a representation of the triple as in the following examples, where the human speaker is called "Lenka":

\begin{enumerate}
    \item I have three white cats: lenka have three-white-cats
    \item can I make a cake: lenka make a-cake
    \item who is from Mexico: ? be-from Mexico
    \item what do you enjoy: agent enjoy ?
\end{enumerate}

Example 1 is a statement, which posts RDF statements to the Knowledge Graph. Example 2 is treated as a verb-question, posted as a SPARQL query to retrieve an answer (yes/no). Examples 3 and 4 are WH-questions with one variable in the triple, also to be converted to SPARQL queries resulting in a(possibly empty) list of answers.

In total, 88 statements and 63 verb-questions are created with aligned triples, and 66 WH-questions were created with at least one variable. We also created a specific series of 86 turns that focus on statements about activities, feelings, kinship and friends, locations, and professions. Finally, we created 28 turns in which perspectives are expressed in relation to social facts. In that case, we add the perspective values for certainty, polarity, and sentiment as scored values in the respective order, as shown below:

\begin{itemize}
    \item john doesn't hate fashion: john hate fashion: 1 -1 -1
    \item I think that birds like flying: birds like flying: 0.75 1 0.75
    \item Bob might be from england: bob be-from england 0.5 1 0
\end{itemize}

In total, we generated 331 test items divided over nine subsets for turn-level evaluation. More examples are given in the Appendix \ref{app:turn-test-example}.

\subsection{Test suite of conversations}
Most triple extraction solutions only consider single sentences or, in the case of conversation, the information expressed in a single turn. However, as explained above, conversations are often more complex than written text such as Wikipedia or news, while information is also exchanged over multiple turns. We therefore annotated a dataset with triples that span sequences of three turns. We sampled data from three complementary sources: PersonaChat \cite{zhang2018personalizing}, DailyDialog \cite{li2017dailydialog} and Circa \cite{louis2020d}. PersonaChat contains approximately 10.907 dialogues and 162.000 utterances obtained through Amazon Mechanical Turk. The crowd workers were randomly paired and assigned a persona, that is, a description of a fake identity, including lines such as 'I am an artist' or 'I live in Amsterdam'. Using a chat interface, workers were tasked to hold a natural conversation with their dialogue partner, whilst conditioning their responses on their assigned persona. DailyDialog is a multiturn open-domain dialogue corpus of spoken English covering a broad range of topics. The dialogues were mined from websites for English language learning. Circa is a data set of polar questions with associated implicit responses collected using crowd annotations. The crowdworkers were asked to imagine questions for ten predefined social situations, which were subsequently posed to other crowdworkers to respond, along with a binary label indicating whether their response confirms or denies the implied statement.






We randomly selected 941 short dialogue fragments of three consecutive utterances: 442 dialogues from Circa, 374 dialogues from DailyDialog and 301 from PersonaChat. The utterances approximately have a length of 29 tokens, excluding end-of-turn markers. Three annotators (authors) annotated the dialogues for social facts by marking the subject, predicate, and object tokens, representing triples claimed by one of the interlocutors. In addition, the annotators could mark any tokens expressing the polarity and certainty of the interlocutor making the claim. Table \ref{tab:conversational-data-annotations} shows an overview of the data and annotations.

\begin{table}[!ht]
    \centering
    \begin{tabular}{|l|l|l|l|}
    \hline
        Total Dialogues & 1,117 & Annotators & 3 \\ \hline
        Turns per Dialogue & 3 & Total Triples & 4,786 \\ \hline
        Total Utterances & 3,351 & ~ & ~ \\ \hline
        Total Tokens & 32,877 & ~ & ~ \\ \hline
        Tokens per Dialogue & 29.43 & Triples per Dialogue & 4.28 \\ \hline
        Tokens per Utterance & 9.81 & Tripples per Utterance & 1.38 \\ \hline
    \end{tabular}
    \caption{Overview of the conversational dataset with triple annotations}
    \label{tab:conversational-data-annotations}
\end{table}

The 941 dialogues contained 2,823 utterances, which received a total of 3,899 annotations of triples and perspectives. We calculated the Inter-Annotator Agreement (IAA) on a pilot set in terms of the Jaccard index and average F measure across the annotators, averaging over all annotator pairs and dialogues in the dataset. Annotation agreement was measured at two levels; the level of triple elements (that is, the sets of subjects, predicates, and objects marked in a dialogue) and the level of triples, as shown in Table \ref{tab:iaa}. As can be seen, we achieved a respectable agreement of 0.688 and 0.773 on the identification of whole triples by the Jaccard and F-measure, respectively. When analyzing the triple elements separately, we do see a slight drop in performance on predicates and objects. However, the quality of the annotations was found to be satisfactory within the time constraints of the annotation effort. 

\begin{table}[!ht]
    \centering
    \begin{tabular}{|l|l|l|}
    \hline
        & Jaccard Index (\%) &  Pairwise F-measure (\%) \\ \hline
        Subjects & 0.871 & 0.916 \\ \hline
        Predicates & 0.79 & 0.872 \\ \hline
        Objects & 0.823 & 0.894 \\ \hline
        Triples & 0.688 & 0.773 \\ \hline
    \end{tabular}
    \caption{Inter-annotator agreement on pilot annotation}
    \label{tab:iaa}
\end{table}

The annotated data set is divided into training and test sets with a 9:1 ratio per dialogue, to ensure that overlapping fragments of the same dialogue fall within the same set. We manually selected 179 dialogues (537 utterances) for testing, exhibiting different linguistic phenomena that characterize conversational information exchange: answer ellipsis, coordination, coreference, statements, negated statements, explicit no-answers and explicit yes-answers. Each of these constitutes a separate test set. A few examples are shown next, where three turns are separated by "<eos>" tokens in the order of "speaker1", "speaker2" and "speaker1". The next line gives the triple, optionally followed by perspective labels for polarity and certainty:

\begin{itemize}
    \item dailydialogs-train-012081
    \begin{itemize}
    \item Yes , it's me.<eos>Do you have a cold ?<eos>No. Worse than that. I have a flu. I'm in bed with a fever.
    \item speaker1,have,a cold,negative
    \end{itemize}
    \item circa-train-002436
    \begin{itemize}
    \item alot really, i enjoy going out to eat with family, going to the movies<eos>would you like to go to the movies tonight ?<eos>no, i'm busy
    \item speaker1,like to,go to the movies,negative
    \end{itemize}
    \item personachat-valid-000114
    \begin{itemize}
    \item ow wow now that will be hard task<eos>yeah do you have kids?<eos>no i sadly do not. i never been one to hold a steady relationship
    \item speaker1,have,kids,negative
    \end{itemize}
    \item circa-train-010351
    \begin{itemize}
    \item we could put on some christian edm and party once your taller to celebrate<eos>do you like edm ?<eos>no, i don't know them
    \item speaker1,like,edm,negative
    \end{itemize}
\end{itemize}

\section{Triple extraction models}
\label{sec:models}
We developed different triple extraction models to extract both factual information and perspectives expressed in dialogues. In particular, we tested five triple extraction approaches:
\begin{enumerate}
    \item Tailored Context-Free Grammar (CFG)
    \item spaCy Dependency Parser
    \item Stanford OpenIE tool
    \item Two fine-tuned BERT-based models
    \item Llama 3.2 prompted with few shot learning 
\end{enumerate}

Each of these triple extractors utilizes different NLP techniques to extract subject-predicate-object triples and their associated perspective values. 
Table \ref{table:nlu_overview} shows an overview of the capabilities of each extractor. In the following subsections, we explain in detail each triple extractor.

\begin{table}[!h]
\centering
\begin{tabular}{|p{2.5cm}|p{2.25cm}|p{1.75cm}|p{1.75cm}|p{1.75cm}|p{1.75cm}|}\hline
\textbf{Triple Extractor} & \textbf{Language} & \textbf{Perspective extraction?}  & \textbf{Question extraction?} & \textbf{Contextual?} & \textbf{Requires training?} \\\hline
CFG & English & Yes & Yes & No & No \\\hline
Spacy & Multilingual & No & No & No & No \\\hline
openIE & English & No & No & No & No \\\hline
conversational & Multilingual & Yes & Yes & Yes & Yes \\\hline
Llama & Multilingual & Yes & Yes & Depends on prompt & No \\\hline
\end{tabular}
\caption{Overview of triple extractors.}
\label{table:nlu_overview}
\end{table}

\subsection{Dependency parser}
We used the spaCy dependency parser and simple dependency patterns mapped to semantic relations as a baseline system. Patterns of frequent phrase structures \cite{wang-etal-2022-extracting} with these dependency relations are converted into subject-predicate-object triples. We extract triples for active and passive variants of subject-predicate-object ("John likes cats"), subject-predicate-adjectival-complement ("John is sick"), subject-predicate-prepositional-object ("John goes to school").

\subsection{Context Free Grammar}
We developed a dedicated Context Free Grammar (CFG) specifically targeting phrase structures of simple statements and questions as single turns. The CFG-based extractor is a rule-based system designed to process individual utterances and identify subject-predicate-object triples. It starts by tokenizing the input and replacing contractions with their long-form equivalents, such as converting "can't" to "cannot". This preprocessing ensures consistency in sentence structure. Next, Part-of-Speech (POS) tagging is applied using both NLTK and Stanford taggers. The processed tokens are then passed through a manually designed context-free grammar, which parses the sentence structure to identify triples. The grammar has 12 rewriting rules to define basic phrase structures and is shown in the appendix \ref{app:cfg}.

The output of the CFG parsing is mapped to subject-predicate-object triples following predefined rules. Tokens are further lemmatized using NLTK, and modal verbs are analyzed with a lexicon to capture additional perspective values, such as uncertainty or temporality. Multiword expressions, such as New York or ice-cream, are detected as collocations and treated as single entities.

Finally, from the phrase structure, we extract triples using specific patterns and a lexicon with 1) lexical categories such as pronouns, (auxiliary) verbs, determiners, quantifiers, 2) semantic categories such as activities, professions, and kinship terms, and 3) sentiment. This detailed semantic analysis ensures that each triple is enriched with meaningful contextual information.

\subsection{Conversational analyzer}
We fine-tuned two BERT models (multilingual BERT-based and Albert) with the annotated sequences of three alternating turns between interlocutors described in Section \ref{sec:conversations}. These models are specifically trained for IOB-style triple extraction and are adapted for conversational contexts by incorporating sequences of three utterances: speaker-agent-speaker turn. The IOB extractor generates all candidates for subjects, predicates, and objects from the three utterances and ranks all possible triple combinations using a trained ranker. This approach is particularly suited to conversational settings, capturing the nuances of how people exchange knowledge. Using the contextual capabilities of BERT, it is expected to excel in detecting triples in diverse and dynamic dialogue scenarios.

\subsection{Few-shot Llama prompting for triple extraction.}
We also used an open source generative model Llama3.2 to extract both triples and perspectives at the turn level. We created a specific prompt for generating a JSON output. The prompt is shown in the Appendix \ref{app:Prompt used for Llama model}. We gave examples for statements (4), WH-questions (5) and a verb-question (1) from our turn-level test set for few shot learning.

\subsection{Stanford Open Information Extraction}
For comparison, we also tested a state-of-the-art triple extractor that was trained on Wikipedia data on our test data. The Stanford OpenIE extractor is designed to extract general relationships from text. Unlike traditional IE tools that rely on a broad set of patterns, this approach simplifies the process by using a smaller set of patterns tailored for canonically structured sentences~\cite{angeli-etal-2015-leveraging} .

To handle longer and more complex sentences, the extractor employs a classifier that isolates self-contained clauses. These clauses are then processed using natural logic inference to identify the most specific arguments for each candidate triple. This approach ensures that the extracted triples are both precise and contextually relevant.

Although the model is trained on Wikipedia, and thus is encyclopedic, it tends to produce a high volume of triples, including noisy or irrelevant ones. Common issues include the extraction of unnecessary modifiers or articles. Post-processing is applied to refine the triples and filter out such noise, improving the overall quality of the output.

\section{Results}
In this section, we present the result on our turn-level and conversation-level test set. Only our conversational models could be applied to the latter. Adapting the Llama prompt for conversational contexts is left for future work.

\subsection{Turn level}
Table \ref{tab:overall-turn-results} shows the results of the extractors over the nine turn-level test sets, where we calculate the precision by taking the number of correct triples divided by correct+incorrect triples.\footnote{Recall is not calculated as there is only one triple per test instance} We consider the precision in extracting the complete triple, the elements of the triples regardless of the combination, and the precision for each element: subject, object and predicate. For most test sets, the CFG model performs best closely followed by the mBERT triple extractor. The latter performed best for triple elements and objects of statements. Remarkably, the Llama model performed best in extracting predicates. OpenIE and the baseline spaCy clearly underperform, which is expected as both have not been created for this genre and specific task. All models suffered mainly from detecting the correct predicate, which is due to the fact that it is not trivial to represent the predicate with a token. Properties are often expressed by longer expressions or even complete texts. Notably, the mBERT model clearly outperformed the Albert model, which may point to multilingual models being more robust for other genres than monolingual models trained on clean texts.




\begin{table}[!ht]
    \centering
    \begin{tabular}{|l|l|l|l|l|l|l|l|l|l|}
    \hline
            test & nr of & model & no  & Precision & Precision & Precision & Precision & Precision \\
         & utter. & & triples & triples & elements & subjects & objects & predicates \\ \hline \hline
        statements & 88 & CFG & 2 & \textbf{51.14} & 67.05 & \textbf{77.27} & 68.18 & \textbf{55.68} \\ \hline
        ~ & ~ & CONV-mBERT & 5 & 45.45 & \textbf{69.32} & 76.14 & \textbf{81.82} & 50.00 \\ \hline
        ~ & ~ & CONV-albert & 27 & 27.27 & 47.73 & 48.86 & 57.95 & 36.36 \\ \hline
        ~ & ~ & LLAMA3.2 & 30 & 12.50 & 34.47 & 45.45 & 36.36 & 21.59 \\ \hline
        ~ & ~ & openIE & 17 & 0.00 & 27.65 & 47.73 & 27.27 & 7.95 \\ \hline
        ~ & ~ & spaCy & 40 & 11.36 & 34.85 & 42.05 & 23.86 & 38.64 \\ \hline
        verb-questions & 63 & CFG & 0 & \textbf{46.03} & \textbf{73.02} & \textbf{87.30} & \textbf{79.37} & \textbf{52.38} \\ \hline
        ~ & ~ & CONV-mBERT & 8 & 38.10 & 61.38 & 65.08 & 69.84 & 49.21 \\ \hline
        ~ & ~ & CONV-albert & 14 & 38.10 & 53.97 & 57.14 & 55.56 & 49.21 \\ \hline
        ~ & ~ & LLAMA3.2 & 5 & 12.70 & 43.92 & 34.92 & 50.79 & 46.03 \\ \hline
        ~ & ~ & openIE & 15 & 0.00 & 26.46 & 57.14 & 22.22 & 0.00 \\ \hline
        ~ & ~ & spaCy & 24 & 15.87 & 41.80 & 55.56 & 25.40 & 44.44 \\ \hline
        wh-questions & 66 & CFG & 2 &\textbf{ 31.82} & \textbf{59.09} & \textbf{57.58} & \textbf{87.88} & 31.82 \\ \hline
        ~ & ~ & CONV-mBERT & 19 & 16.67 & 37.88 & 34.85 & 60.61 & 18.18 \\ \hline
        ~ & ~ & CONV-albert & 3 & 18.18 & 49.49 & 50.00 & 66.67 & 31.82 \\ \hline
        ~ & ~ & LLAMA3.2 & 0 & 15.15 & 51.52 & 43.94 & 74.24 & \textbf{36.36} \\ \hline
        ~ & ~ & openIE & 58 & 0.00 & 4.04 & 7.58 & 4.55 & 0.00 \\ \hline
        ~ & ~ & spaCy & 64 & 3.03 & 3.03 & 3.03 & 3.03 & 3.03 \\ \hline \\
        \end{tabular}
            \caption{Results of the triple extraction on the utterance level test set for four models: CFG=Context Free Grammar, CONV-mBERT= multilingual BERT-base fine-tuned with conversational triples, CONV-Albert=Albert-base finetuned with conversaitons, Llama3.2=Llama3.2 prompted with few-shots, openIE=Stanford's open information extraction, spaCy=spaCy dependency patterns. Three different test sets are given for statements, verb-questions and wh-questions. Precision is calculated for predicted complete triples, the triple elements combined and the separate subjects, predicates and objects by correct predictions divided by correct \& wrong-predictions. The no triples column shows the number of utterances that did not receive a triple.}
    \label{tab:overall-turn-results}
\end{table}

When we consider the specific domains of the statements, we see in Table \ref{tab:specific-turn-results} that the CFG and mBERT models score more similar and both, again, mostly outperform the others. Exceptions are kinship-friendship relations where Llama and openIE outperform the others, and spaCy performs best on activity triple elements. Again, we see that all models struggle most with predicting the tokens for the predicates.

\begin{table}[!ht]
\small{
    \centering
    \begin{tabular}{|l|l|l|l|l|l|l|l|l|l|}
    \hline
        test & nr of & Model & no  & Precision  & Precision & Precision & Precision & Precision \\
        & utterances & & triples & triples  & elements & subjects & objects & predicates \\ \hline \hline
        activities & 24 & CFG & 0 & 0.00 & 41.67 & 91.67 & 33.33 & 0.00 \\ 
        ~ & ~ & CONV-mBERT & 1 & 0.00 & 37.50 & 79.17 & 33.33 & 0.00 \\ 
        ~ & ~ & CONV-albert & 4 & 0.00 & 30.56 & 70.83 & 20.83 & 0.00 \\ 
        ~ & ~ & LLAMA3.2 & 10 & 0.00 & 19.44 & 54.17 & 4.17 & 0.00 \\ 
        ~ & ~ & openIE & 5 & 0.00 & 29.17 & 25.00 & 62.50 & 0.00 \\ 
        ~ & ~ & spaCy & 5 & 0.00 & 50.00 & 75.00 & 75.00 & 0.00 \\ \hline
        feelings & 8 & CFG & 0 & 50.00 & 83.33 & 100.00 & 100.00 & 50.00 \\ 
        ~ & ~ & CONV-mBERT & 0 & 50.00 & 83.33 & 100.00 & 100.00 & 50.00 \\ 
        ~ & ~ & CONV-albert & 0 & 50.00 & 83.33 & 100.00 & 100.00 & 50.00 \\ 
        ~ & ~ & LLAMA3.2 & 3 & 12.50 & 29.17 & 62.50 & 12.50 & 12.50 \\ 
        ~ & ~ & openIE & 0 & 0.00 & 58.33 & 50.00 & 100.00 & 25.00 \\ 
        ~ & ~ & spaCy & 0 & 50.00 & 83.33 & 100.00 & 100.00 & 50.00 \\ \hline
        kinship-friends & 34 & CFG & 2 & 0.00 & 10.78 & 29.41 & 2.94 & 0.00 \\ 
        ~ & ~ & CONV-mBERT & 1 & 0.00 & 10.78 & 29.41 & 2.94 & 0.00 \\ 
        ~ & ~ & CONV-albert & 3 & 0.00 & 7.84 & 20.59 & 2.94 & 0.00 \\ 
        ~ & ~ & LLAMA3.2 & 6 & 0.00 & 14.71 & 41.18 & 2.94 & 0.00 \\ 
        ~ & ~ & openIE & 0 & 0.00 & 12.75 & 35.29 & 2.94 & 0.00 \\ 
        ~ & ~ & spaCy & 8 & 0.00 & 9.80 & 29.41 & 0.00 & 0.00 \\ \hline
        locations & 13 & CFG & 0 & 100.00 & 100.00 & 100.00 & 100.00 & 100.00 \\ 
        ~ & ~ & CONV-mBERT & 0 & 0.00 & 58.97 & 100.00 & 76.92 & 0.00 \\ 
        ~ & ~ & CONV-albert & 1 & 0.00 & 53.85 & 92.31 & 69.23 & 0.00 \\ 
        ~ & ~ & LLAMA3.2 & 11 & 0.00 & 10.26 & 15.38 & 15.38 & 0.00 \\ 
        ~ & ~ & openIE & 0 & 0.00 & 23.08 & 69.23 & 0.00 & 0.00 \\ 
        ~ & ~ & spaCy & 0 & 0.00 & 41.03 & 100.00 & 0.00 & 23.08 \\ \hline
        professions & 7 & CFG & 0 & 71.43 & 85.71 & 100.00 & 85.71 & 71.43 \\ 
        ~ & ~ & CONV-mBERT & 0 & 57.14 & 71.43 & 100.00 & 57.14 & 57.14 \\ 
        ~ & ~ & CONV-albert & 1 & 42.86 & 57.14 & 85.71 & 42.86 & 42.86 \\ 
        ~ & ~ & LLAMA3.2 & 4 & 0.00 & 23.81 & 42.86 & 14.29 & 14.29 \\ 
        ~ & ~ & openIE & 0 & 0.00 & 42.86 & 42.86 & 85.71 & 0.00 \\ 
        ~ & ~ & spaCy & 0 & 28.57 & 71.43 & 100.00 & 85.71 & 28.57 \\ \hline
    \end{tabular}}
        \caption{Results of the triple extraction on different test sets for activities, feelings, kinship-friend relations, locations, and professions. Results for four models: CFG=Context Free Grammar, CONV-mBERT= multilingual BERT-base fin-tuned with conversational triples, CONV-Albert=Albert-base fine-tuned with conversations, Llama3.2=Llama3.2 prompted with few-shots, openIE=Stanford's open information extraction, spaCy=spaCy dependency patterns.  Precision is calculated for predicted complete triples, the triple elements combined and the separate subjects, predicates and objects by correct predictions divided by correct \& wrong-predictions. The no triples column shows the number of utterances that did not receive a triple.}
    \label{tab:specific-turn-results}
\end{table}

This two-sided evaluation highlights a key trade-off between the systems. Structure-based approaches, such as CFG and spaCy, are effective in extracting individual triple elements but struggle to assemble correct complete triples and specifically the predicates. Neural models like BERT and Llama, on the contrary, do a better job on complex relations such as the predicates but fail to get the complete triple correct. Finally, we observed that the CFG extractor significantly outperformed the other models to handle questions, whether verb or WH-questions. This suggests that rule-based approaches are particularly suited to the structured nature of interrogative sentences.

We further tested the models on statements that also explicitly express perspectives on social facts such as polarity, certainty, and sentiment. The results are shown in Table \ref{tab:perspective-results}. Statements with perspectives have values for both the triples and the perspectives. 

\begin{table}[!ht]
    \centering
    \begin{tabular}{|l|l|l|l|l|l|l|l|}
    \hline
        Model & no  & Precision  & Precision & Precision & Precision & Precision & Precision \\
        & triples & triples  & elements & subjects & objects & predicates & perspective \\ \hline \hline
        CFG & 0 & 32.14 & 45.24 & 57.14 & 46.43 & 32.14 & \textbf{66.67} \\ 
        CONV-mBERT & 4 & \textbf{35.71} & \textbf{58.33} & \textbf{60.71} & \textbf{78.57} & \textbf{35.71} & 51.85 \\ 
        CONV-albert & 15 & 21.43 & 34.52 & 35.71 & 46.43 & 21.43 & 60.00 \\ 
        LLAMA3.2 & 17 & 10.71 & 23.81 & 28.57 & 25.00 & 17.86 & ~ \\ 
        openIE & 19 & 0.00 & 15.48 & 28.57 & 17.86 & 0.00 & ~ \\ 
        spaCy & 13 & 35.71 & 45.24 & 53.57 & 39.29 & 42.86 & ~ \\ \hline
    \end{tabular}
    \caption{Results of the triple extraction on dedicated test set (28 utterances) expressing various perspectives (certainty, denial, sentiment) of speakers on the claims in utterances. Results for four models: CFG=Context Free Grammar, CONV-mBERT= multilingual BERT-base fine-tuned with conversational triples, CONV-Albert=Albert-base fine-tuned with conversations, Llama3.2=Llama3.2 prompted with few-shots, openIE=Stanford's open information extraction, spaCy=spaCy dependency patterns.  Precision is calculated for predicted complete triples, the triple elements combined and the separate subjects, predicates and objects by correct predictions divided by correct \& wrong-predictions. The no triples column shows the number of utterances that did not receive a triple.}
    \label{tab:perspective-results}
\end{table}

We expect the extraction of the triples to be more complicated because the linguistic structure reflects different layers of information. In general, we see that the results for the triples are, in fact, lower for all models compared to the results for the statements in Table \ref{tab:overall-turn-results}. Remarkably, spaCy is the only exception and performs better compared to plain statements. Apparently, the dependency patterns are more robust for other phrases such as adverbs or adjectives that enrich the statements with perspectives. Across the models, the mBERT conversational model performs the best for triples. However, CFG performs best when it comes to the specific perspective values. It is also interesting that the Albert conversational model outperforms mBERT for perspective values. This may be explained by its sensitivity to discourse relations due to the next-sentence-order learning objective. The LLama, openIE and spaCy models did not provide any output for the perspective values as they were not designed to do so.

\subsection{Conversation level}

Only the conversational models were designed to extract triples from contextual sequences of turns. Table \ref{tab:conv-results} shows the results for the best model mBERT, where the test data is differentiated for separate linguistic phenomena.

\begin{table}[!ht]
\small{
\centering
    \begin{tabular}{|l|l|l|l|l|l|l|l|l|l|}
    \hline
        test & nr. of & no & Precision & precision  & precision & precision & precision & precision \\
         & utterances & triples & triples  & elements & subjects & objects & predicates & perspectives \\ \hline \hline  answer\_ellipsis & 101 & 4 & 1.98  & 28.05 & 61.39 & 10.89 & 11.88 & 66.67 \\ \hline
        coordination & 82 & 1 & 6.10  & 46.75 & 76.83 & 32.93 & 30.49 & 66.67 \\ \hline
        coreference & 78 & 0 & 1.28  & 24.79 & 29.49 & 35.90 & 8.97 & 66.67 \\ \hline
        statements & 91 & 1 & 2.20  & 37.36 & 76.92 & 24.18 & 10.99 & 66.67 \\ \hline
        negated statements & 79 & 1 & 3.80 & 45.15 & 79.75 & 40.51 & 15.19 & 66.24 \\ \hline
        explicit no answers & 47 & 0 & 4.26  & 43.26 & 78.72 & 25.53 & 25.53 & 65.96 \\ \hline
        explicit yes answers & 59 & 0 & 3.39  & 36.16 & 74.58 & 11.86 & 22.03 & 66.67 \\ \hline \hline
        Total & 537 & 7 & 3.29  & 37.36 & 68.24 & 25.97 & 17.87 & 66.50 \\ \hline
    \end{tabular}}
    \label{tab:conv-results}
    \caption{Results of the conversation level evaluation for the conversational model. The test is divided over different conversational phenomena such as ellipsis, coordination, coreference, statements, negations of statements, no answers and yes answers. The no triples column shows the number of utterances that did not receive a triple.}
\end{table}

Overall, the results are lower than for the turn-level evaluation, which is expected, as this is a more challenging task. The precision at the complete triple level is very low, and the precision for predicates is again the lowest from the elements. The subject elements score highest in most cases, except for conversations exhibiting coreference. Most of these subjects are expected to refer to the interlocutors, which are more easily detectable by mapping the pronouns "I" and "you" correctly. Finally, the extraction of perspectives scores fairly well overall compared to the result in Table \ref{tab:perspective-results}. This is in part due to the fact that only negative polarity and uncertainty were annotated and tested, which are both often explicitly marked.

\section{Conclusion}
\label{sec:conslusion}
Triple extraction from social conversations is more difficult than extraction of information from news or Wikipedia texts, which is studied more extensively. First, information is exchanged differently and in a more complex way in conversation, spread over different turns across interlocutors, through incomplete sentences that exhibit ellipsis, coreference, coordination, and implicit statements. Second, the information itself is more complex in social exchange compared to news and Wikipedia facts. It involves a wider and more open range of properties and values that address social relationships, personal feelings, and history.

 We released two data sets for training and testing the triple extraction of social conversation. We also developed several models, among which two models for the difficult task of extracting social facts and perspectives from ongoing conversations involving sequences of turns. We hope that this enables future research to model social conversations more explicitly.

\section{Limitations}
The turn-level evaluation is based on an artificial data set with simple utterances. The Context-Free Grammar (CFG) was optimized for this test set, whereas the other models were not. The CFG should therefore be seen as an upper ceiling of performance, whereas the spaCy model is a baseline. The conversational data are more natural, but are created by crowd workers having a conversation among them, which is different from people having a conversation with an artificial agent. This makes the conversation-level evaluation less representative, and the performance is also lower than expected for the human-agent interaction, where humans will adapt to systems. Both evaluation data sets are written chat data. In the case of spoken data, we expect the performance to be lower. Only the CFG and conversational models were created to extract perspective values in addition to the triples, but both were not optimized for perspective value detection either.

We have not used any prompt engineering for the Llama model. Fine-tuning and prompt engineering could greatly improve Llama's results. We also did not provide the Llama model with the previous dialog history to extract triples from the current utterance. In turn-level tests, no dialog context is provided. We could also prompt Llama for conversational data as in the conversation-level test.

So far we tested the models only on English. The Llama and the multilingual BERT model could also be applied to test data in other languages. In future work, we plan to extend the evaluation and training data to other languages.

\section{Ethics}
The data used in the conversation-level test are created by crowd-workers. This may have created some bias in the data. Furthermore, all the data are acted on and are not supposed to contain true personal information. However, there is no guarantee that crowdworkers did not communicate personal information. Since the data are publicly available, we did not consider it necessary to consult an ethics committee.

\section*{Acknowledgments}
This research was funded by the Vrije Universiteit Amsterdam and the Netherlands Organisation for Scientific Research (NWO) through the \textit{Hybrid Intelligence Centre} via the Zwaartekracht grant (024.004.022), and the \textit{Spinoza} grant (SPI 63-260) awarded to Piek Vossen. 

\bibliographystyle{unsrt}  
\bibliography{main}  

\section{Appendix}

\subsection{Turn-level evaluation data examples}
\label{app:turn-test-example}

\begin{itemize}
    \item Statements
\begin{itemize}
    \item I know you: lenka know agent
    \item where is she: she be ?
    \item my best friend is he: lenka best-friend-is he
    \item I have three white cats: lenka have three-white-cats
    \item I think Selene doesn't like cheese: selene like cheese
    \item I think Selene hates cheese: selene hate cheese
    \item selene might come today: selene might-come today
    \item I don't think selene likes cheese: selene like cheese
\end{itemize}
\end{itemize}

\begin{itemize}
    \item Verb-questions:
\begin{itemize}
    \item can a bird sing a song: a-bird can-sing a-song
    \item can I call you: Lenka can-call agent
    \item can I make a cake: Lenka can-make a-cake
    \item will you go to Paris: agent will-go-to paris
    \item must you go home: agent must-go home
    \item do you like amsterdam: agent like amsterdam
\end{itemize}
\end{itemize}

\begin{itemize}
    \item Wh-Questions
\begin{itemize}
    \item who is your best friend: agent best-friend-is ?
    \item who can sing: ? can-sing ?
    \item who likes talking to people: ? like talking-to-people
    \item who works at the university: ? work-at the-university
    \item who have you seen: agent see ?
    \item who is from Mexico: ? be-from Mexico
    \item who does Selene know: Selene know ?
    \item what is your favorite color: agent favorite-color-is ?
    \item what is my favorite food: Lenka favorite-food-is ?
    \item what do you enjoy: agent enjoy ?
    \item where is selene from: Selene be-from ?
    \item where is your friend: agent friend-is ?
\end{itemize}
\end{itemize}

\begin{itemize}
    \item Perspectives
\begin{itemize}
    \item john doesn't hate fashion: john hate fashion: 1 -1 -1
    \item I think selene works in Amsterdam: selene work-in amsterdam: 0.75 1 0
    \item I think john can't come to school: john can-come-to school: 0.75 -1 0
    \item you know I like coffee: lenka like coffee: 1 1 0.75
    \item I think that birds like flying: birds like flying: 0.75 1 0.75
    \item I know that you are not a human: agent be a-human: 1 -1 0
    \item you must bring three books: agent must-bring three-books 1 1 0
    \item john might like reading books: john might-like reading-books 0.5 1. 0.75
    \item selene should come to the university: selene should-come-to the-university 0.75 1 0
    \item Bob might be from england: bob might-be-from england 0.5 1 0
\end{itemize}
\end{itemize}

\subsection{Context-Free Grammar rules}
\label{app:cfg}

\begin{verbatim}
U -> S | Q
Q -> W VP C| W NP C | W VP NP C | W VP NP VP C | W NP VP C | V NP C | V NP C C
W -> WRB | WP | WDT
S -> NP VP C | NP C | NP VP S
C -> NP | VP | PREP
NP -> N | JJ | J NP | D NP | N NP | RB NP | PREP NP
VP -> V VP | V RB | RB VP | V | V PREP VP | V PREP
PREP -> IN | TO
V -> VBD | VBP | VBZ | VBN | VBG | VB | MD
D -> DT | CD | PRPPOS
N -> NN | NNS | NNP | NNPS | PRP | DT
J -> JJ | JJR | JJS
\end{verbatim}

\subsection{Llama prompt and few shot examples}
\label{app:Prompt used for Llama model}

\begin{verbatim}
_INSTRUCT = {'role':'system', 'content':'You will analyze a dialogue and 
break it down into triples consisting of a subject, predicate,and object.
Each triple should capture the essence of interactions between speakers.
Replace the predicate by its lemma, for example "is" and "am" should become "be".
Remove auxiliary verbs such as "be", "have", "can", "might" from predicates.
If the object starts with a preposition, concatenate the preposition to the 
predicate separated by a hyphen, for example "be-from".
Additionally, annotate each triple with:
- Sentiment (-1 for negative, 0 for neutral, 1 for positive)
- Polarity (-1 for negation, 0 for neutral/questioning, 1 for affirmation)
- Certainty (a scale between 0 for uncertain and 1 for certain)
Ensure that predicates are semantically meaningful. 
Separate multi-word items with an underscore. 
Save it as a JSON with this format: 
{"dialogue": [{"sender": "human", "text": "I am from Amsterdam.", 
"triples": [ { "subject": "I", "predicate": "be_from", "object": "Amsterdam", 
"sentiment": 0, "polarity": 1, "certainty": 1}]},
{"dialogue": [{"sender": "human", "text": "lana is reading a book.", 
"triples": [ { "subject": "lana", "predicate": "read", "object": "a-book", 
"sentiment": 0, "polarity": 1, "certainty": 1}]},
{"dialogue": [{"sender": "human", "text": "You hate dogs.", 
"triples": [{ "subject": "You", "predicate": "hate", "object": "dogs", 
"sentiment": -1, "polarity": 1, "certainty": 0.7}]},
{"dialogue": [{"sender": "human", "text": "Selene does not like cheese.", 
"triples": [ { "subject": "Selene", "predicate": "like", "object": "cheese", 
"sentiment": -1, "polarity": -1, "certainty": 0.5}]},
{"sender": "human","text": " Who likes cats?", 
"triples": [ {"subject": "", "predicate": "like", "object": "cats", 
"sentiment": 1, "polarity": 1, "certainty": 0.1}]},
{"sender": "human","text": " Wen did Selene come?", 
"triples": [ {"subject": "Selene", "predicate": "come", "object": "", 
"sentiment": 1, "polarity": 1, "certainty": 0.1}]},
{"sender": "human","text": " Where can I go?", 
"triples": [ {"subject": "I", "predicate": "go", "object": "", 
"sentiment": 1, "polarity": 1, "certainty": 0.1}]},
{"sender": "human","text": " Who likes cats?", 
"triples": [ {"subject": "", "predicate": "like", "object": "cats", 
"sentiment": 1, "polarity": 1, "certainty": 0.1}]},
{"sender": "human","text": " Are cats pets?", 
"triples": [ {"subject": "cats", "predicate": "be", "object": "pets", 
"sentiment": 1, "polarity": 1, "certainty": 0.1}]}]}
{"sender": "human","text": " Do you like cats?", 
"triples": [ {"subject": "you", "predicate": "like", "object": "cats", 
"sentiment": 1, "polarity": 1, "certainty": 0.1}]}]}
Do not output any other text than the JSON.'
}
    
\end{verbatim}

\subsection{Predicates detected by the conversational triple extractor}
\label{app:conv_predicates}

\begin{table}[!ht]
\scriptsize{
    \centering
    \begin{tabular}{|l|l|l|l|l|l|l|l|l|}
    \hline
        act & become & cut & find & know & occupy & recycle & socialize & waste \\ \hline
        add & believe & dance & finish & lead & open & relationship & sort & watch \\ \hline
        afford & betray & decide & fit & learn & order & relax & speak & water \\ \hline
        agree & better than & depend on & fix & leave & owe & remain & spend & wear \\ \hline
        aid & borrow & destroy & follow & light & own & remember & start & will \\ \hline
        allow & break & devote & forget & like & paint & remove & starve & win \\ \hline
        anticipate & break up & die & gain & limit & park & reserve & stay & work \\ \hline
        arrest & bring & different & get & listen & pay & respond & steal & work with \\ \hline
        arrive & burn & dislike & get in & live & pay attention & retire & stop & worry \\ \hline
        ask & buy & distrust & give & live in & pick & return & study & would \\ \hline
        be & call & do & go & located & plan & rule & survive & write \\ \hline
        be angry & can & do badly & grow & lock out & play & run & swim & ~ \\ \hline
        be arrested & cancel & do well & handle & look & portray & save & take & ~ \\ \hline
        be at & care & draw & harder than & lose & practice & say & take off & ~ \\ \hline
        be available & catch & dress & have & love & pray & see & take out & ~ \\ \hline
        be certain & cause & drink & hear & made of & prefer & seek & talk & ~ \\ \hline
        be difficult & change & drive & help & make & prepare & seem & taste & ~ \\ \hline
        be free & check in & drop & hit & marry & pretend & sell & teach & ~ \\ \hline
        be from & choose & earn & hold & match & prevent & send & tell & ~ \\ \hline
        be happy & clean & eat & hope & mean & promise & share & think & ~ \\ \hline
        be in & close & enable & hurry & meet & propose & shoot & throw & ~ \\ \hline
        be in between & come & endure & hurt & miss & protect & should & tired & ~ \\ \hline
        be old & confuse & equal & include & misspell & pull & show & travel & ~ \\ \hline
        be on & consider & exchange & investigate & mix & put & sign & try & ~ \\ \hline
        be out & contact & exercise & involve & more expensive & rain & similar & use & ~ \\ \hline
        be over & cook & expect & join & motivate & raise & sing & visit & ~ \\ \hline
        be ready & copy & fall & joke & move & reach & sit & wait & ~ \\ \hline
        be scared & cost & feel & keep & must & read & sleep & wake up & ~ \\ \hline
        be with & could & fight & keep secret & need & recognize & smell & walk & ~ \\ \hline
        be wrong & count & fill & kill & None & recommend & smoke & want & ~ \\ \hline
    \end{tabular}}
\end{table}

\end{document}